\pdfoutput=1

\documentclass[11pt]{article}

\usepackage[]{acl}

\usepackage{times}
\usepackage{latexsym}

\usepackage{times}
\usepackage{url}
\usepackage{latexsym}
\usepackage{natbib}
\usepackage{layout}
\usepackage{float}
\usepackage{caption}
\usepackage{graphicx}

\usepackage[T1]{fontenc}

\usepackage[utf8]{inputenc}

\usepackage{microtype}

\usepackage{tabularx}

\title{Domain Specific Sub-network for Multi-Domain Neural Machine Translation}

\author{Amr Hendy, Mohamed Abdelghaffar, Mohamed Afify and Ahmed Y. Tawfik
\\
Microsoft Egypt Development Center, Cairo, Egypt
\\ 
\texttt{\{amrhendy,mohamed.abdelghaar,mafify,atawfik\}@microsoft.com}
}

\begin{document}
\maketitle
\begin{abstract}
This paper presents Domain-Specific Sub-network (DoSS).
It uses a set of masks obtained through pruning to define a sub-network for each domain and finetunes the sub-network parameters on domain data. This performs very closely and drastically reduces the number of parameters compared to finetuning the whole network on each domain.
Also a method to make masks unique per domain is proposed and shown to greatly improve the generalization to unseen domains. In our experiments on German to English machine translation the proposed method outperforms the strong baseline of continue training on multi-domain (medical, tech and religion) data by 1.47 BLEU points. Also continue training DoSS on  new domain (legal) outperforms the multi-domain (medical, tech, religion, legal) baseline by 1.52 BLEU points.
\end{abstract}

\section{Introduction}
Neural machine translation (NMT) has witnessed significant advances based on transformer models \cite{vaswani2017attention}. These models are typically trained on large amounts of data from different sources, i.e. general 
data, from a single language pair or multiple languages \cite{aharoni-etal-2019-massively}.  
The fact that the models are trained on general data usually leads to poor, or less than average,  performance on specific domains. This has a lot of practical implication since many 
users of machine translation are interested in the performance 
on some specific domain(s). Therefore, improving the performance of NMT on specific domains has become an active area of 
research. We refer the reader to \cite{chu-wang-2018-survey} for a review. Broadly speaking, the proposed techniques could be divided into data-centric and model-centric approaches. The goal of the former methods is to acquire, often automatically, monolingual and bilingual data that is representative of the domain of interest. The latter techniques, on the other hand, focus on modifying the model to perform well on the domain of interest without sacrificing the performance on general data.

Finetuning of the model parameters using domain data is perhaps one of the earliest and most popular techniques for domain adaptation \cite{FreitagA16}. Parallel domain data is usually limited and to avoid overfitting  different techniques as model interpolation \cite{MItchell2022Wise}, regularization \cite{miceli-barone-etal-2017-regularization} and mixing domain and general data \cite{chu-etal-2017-empirical} are used. Also other methods that introduce additional parameters in a controllable way have been successfully introduced such as adapters \cite{bapna-firat-2019-simple} and low-rank adaptation (LoRA) \cite{Hu2021LoRA}.

In \cite{FrankleCarbin2018lotteryticket} it is shown that identifying sub-networks by pruning a large network, referred to as winning tickets, and retraining them leads to equal accuracy to the original network. This idea is explored for multilingual neural machine translation (MNMT) using the so-called language specific sub-networks (LaSS) \cite{lin-etal-2021-learning}. Here we further explore the idea for domain finetuning and refer to it as Domain Specific Sub-network (DoSS). The basic idea is to identify a sub-network per domain via pruning and masking. The sub-network has both shared parameters with other domains as well as domain-specific parameters. It should be noted that the mask 
can overlap for multiple domains which results in some 
parameters shared by multiple domains. We also explore using constrained masks where we ensure that each mask represents only one domain. The latter is expected to work better for adding unseen domains. In contrast to language, domain information may not be necessarily known at inference time. In this work, similar to common domain fientuning setups, we assume the domain information is known but using a domain classifier at runtime should be straight forward. Given the domain information, inference can be carried with the trained model and the domain mask.

The paper is organized as follows. Section \ref{method} gives a detailed description of the proposed method followed by the experimental results in Section \ref{experiments}. Finally, the conclusion is given in Section \ref{conclusion}. 

\begin{figure*}[!htp]
    \centering
    \includegraphics[scale=0.75]{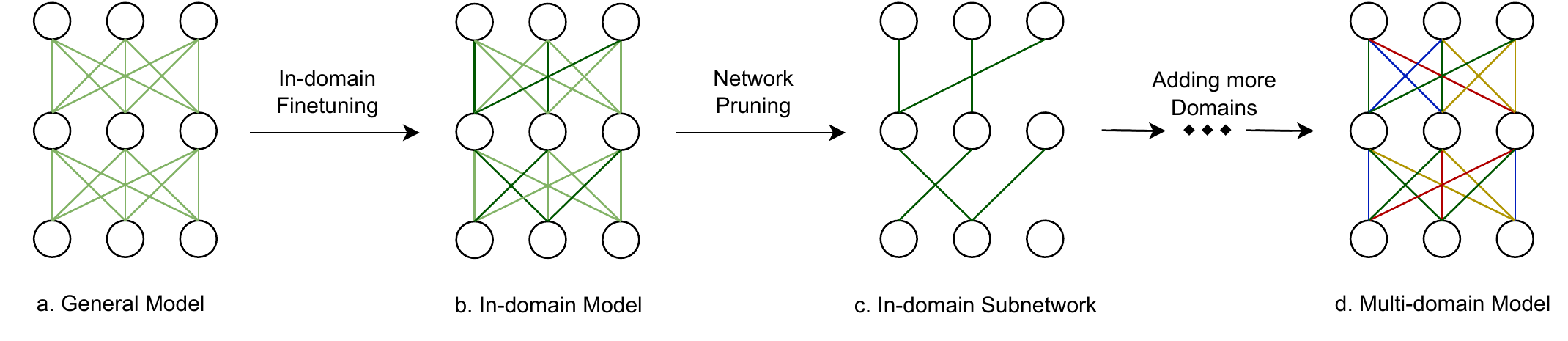}
    \caption{Illustration of domain adaptation from the general domain to the multi-domain setup with DoSS.}
    \label{fig:doss-diagram}
\end{figure*}

\section{Method}
\label{method}

We present the DoSS method in this section as shown in Figure \ref{fig:doss-diagram}. We focus on the bilingual setting and defer the multilingual case to future work. Assume we have an initial model $\lambda_{0}$ that is trained on large amounts of general data. We also have the data sets $\{{\cal D}_{i}\}_{i=1}^{N}$ corresponding to $N$ domains and each data set consists of $L_{i}$ sentence pairs $(x_{j},y_{j})$.Typically, the initial model is finetuned for each domain resulting in $N$ domain models. Here, we first create a mask for each domain using pruning then train a domain sub-network using the resulting masks. We will explain the two steps below.
\subsection{Creating Domain Masks}
\label{masks}
We create a binary mask ${\bf M}_{i}$for each domain that has a 0 or 1 for each model parameter. Following \cite{lin-etal-2021-learning} we calculate the domain masks as follows:
\begin{enumerate}
    \item  Start from initial model $\lambda_{0}$.
    \item For each domain $i$ finetune $\lambda_{0}$ using the corresponding domain data ${\cal D}_{i}$ for $[5:10]$ epochs. This will intuitively amplify the important weights for the domain and diminish other weights. This finetuning stage requires only a few epochs compared to the full finetuning training budget that makes it an effective way to build the mask.

    \item Sort the weights of the finetuned model and prune the lowest $\alpha$ in the encoder and the lowest $\beta$ in the decoder. We found that using separate pruning parameters for the encoder and the decoder gives us better control on the resulting sub-networks. The mask for domain $i$ is created by setting the upper $1 - \alpha$ percent in the encoder and $1 - \beta$ percent in the decoder to 1 and all other elements to 0. 
\end{enumerate}
The above mask creation algorithm is unconstrained in the sense that multiple domains can share the same weight. This has no problem as long as we train multiple domains simultaneously as given below but will degrade performance if we want to add a new domain after the model has been trained for a set of domains. Therefore, we experiment here with simple constrained mask creation where step 3 is modified to set a mask element to 1 if it belongs to the top $1- \alpha (\beta)$ percent in the encoder (decoder) and doesn't belong to other domain masks. This makes the subnetwork parameters unique but is dependent on the order the domains are presented and can cover at most $\min(1/1-\alpha,1/1-\beta)$ domains. Looking into more sophisticated constrained methods could be a topic for future research. Once the domain masks are created we train the sub-networks again following a similar algorithm to \cite{lin-etal-2021-learning}.

\subsection{Training the Sub-networks}
\label{subnetworks}
Here we follow the so-called structure aware joint training. Given the initial model $\lambda_{0}$ and the domain masks 
${\bf M}_{i}$ we finetune the initial model using the domain data. The finetuning is done in a mask-aware manner where the mini-batches are formed per domain $i$ and for each mini-batch we only update parameters where ${M}_{i}$ equals 1. This way we end up with a single model $\lambda$ where shared parameters come from the original model and the domain-specific parameters come from the structure-aware training.

\subsection{Inference}
\label{inference}
Inference is done using the model $\lambda$ and its masks ${\bf M}$. For an input utterance coming from domain $i$ inference is done using the parameters $\lambda \odot {\bf M}_{i}$ where this stands of using the finetuned parameters from the mask and the original parameters otherwise. Domain information is often not known in test time but in this work we assume that the domain is known and perform inference on batches from the same domain for efficiency. When domain is unknown we can use a domain classifier at run-time. We will test this approach in future work.
\section{Experiments and Results}
\label{experiments}
We evaluate the performance of DoSS on German to English translation, and we consider three domains: medicine, religion, and technology. The baseline model was a German to English model trained on 32.13M parallel sentences that were provided by the WMT19 news translation shared task\footnote{\url{https://www.statmt.org/wmt19/translation-task.html}}. All domain and baseline data are filtered to remove sentences longer than 250 tokens, as well as sentences with a source to target length ratio smaller than 0.67 or exceeding 1.5. Fasttext \cite{grave2018learning} language identification was also applied to both sides of the bitext to reduce the garbage \cite{ng2019facebook}.

\subsection{Experimental Setup}

DoSS is implemented as a Fairseq \cite{ott2019fairseq} extension and the model uses a big transformer architecture \cite{vaswani2017attention} with $6$ encoder layers and $6$ decoder layers with $1024$ model dimension and $8192$ feed-forward layer hidden dimension with $16$ attention heads. We use pre-layer normalization which is becoming more standard for the transformer architecture \citep{xiong2020layer}. We use vocabulary of size $42,000$ with the fastBPE tokenizer\footnote{\url{https://github.com/glample/fastBPE}}. The model size is $270$M parameters.

The training uses Adam optimizer and inverse square root learning rate scheduler. All hyper parameters for the domain experiments are given in Table \ref{tab:hyperparams}.  All the models are trained on 8 NVIDIA Tesla V100 GPUs with 32GB memory.
\begin{table}[ht!]
\small
\centering
\label{tab:hyperparams}
\begin{tabular}{l l l l}
    \hline
    Hyper Parameter& Pretraining & Finetuning & DoSS \\
    \hline
    Learning Rate & $0.0005$ & $0.0001$ & $0.0001$ \\
    Warmup        & $4000$  & $1000$ & $1000$ \\
    Batch Size    & $4$k    & $4$k  & $4$k  \\
    Dropout       & $0.1$   & $0.3$ & $0.1$  \\
    \hline
\end{tabular}

\caption{Hyper-parameters comparison between experiment sets.}

\label{tab:hyperparams}

\end{table}

\subsection{Domain Data}

For the domain data collection, we base our collection on \cite{khayrallah2018regularized}.
The medical domain data consists of the German to English corpus of the European Medicines Agency (EMEA). The religion domain data consists of German and English translations of Quran in the Tanzil corpus. For the tech domain we use a joint corpus consisting of Gnome, KDE, PHP, Ubuntu and Open Office. The legal domain data consists of JRC-Acquis data for this language pair. All data obtained from OPUS \cite{TIEDEMANN12.463}. Table \ref{tab:dataSizes} summarizes the data sizes in each domain before and after applying the filtration process described earlier in this section.
\begin{table}[h!]
    \centering
    \small
    \begin{tabular}{l l l}
    \hline
        Corpus & Raw (K) & Filtered (K) \\
        \hline
        WMT & 38,69 & 32,13\\
        EMEA &  1,104 & 647\\
        Tanzil&  480 & 418\\
        JRC Aquis & 715& 637\\
        Tech & 338 & 177\\
        \hline
    \end{tabular}
    \caption{Domain data sizes before and after filtration}
    \label{tab:dataSizes}
\end{table}

\subsection{Domain Finetuning versus DoSS}

We conducted a set of four fine-tuning runs to fine-tune the base model using the data for each domain separately and  one run in which we fine-tuned the base model using the data from all three domains jointly (All-FT). Table \ref{tab:finetuning} shows that generally fine-tuning on the same domain results in a better performance on that particular domain while fine-tuning on all domains jointly represents a reasonable compromise. Moreover, DoSS yields a better model than All-FT by $1.47$ BLEU points and reduces the average difference between domain-specific fine-tuning from 2.04 BLEU points in the case of All-FT to just 0.46 BLEU points. 

\begin{table}[ht!]
    \centering
    \small
    \begin{tabular}{l r r r | r}
    \hline
         & EMEA  & Tech & Tanzil  & Average\\
        \hline
        Baseline &41.52&33.00&16.70&30.41\\
        \hline
        EMEA &53.57 & 22.88 & 9.32 & 28.59 \\
        Tech &28.12 & \textbf{57.71} & 11.11 &	32.31 \\
        Tanzil &2.42& 3.67& 18.79 & 8.29\\
        \hline
        All-FT &53.26 & 52.01 &\textbf{19.01} &41.42\\
        \hline
        DoSS &\textbf{54.03} &56.04 &18.59&\textbf{42.89}\\
    \end{tabular}
    \caption{SacreBLEU scores for domain finetuning experiments. \textbf{Baseline} is the general model trained on WMT19. \textbf{EMEA} is the baseline model finetuned on EMEA domain data. \textbf{Tech} is the baseline model finetuned on Tech domain data. \textbf{Tanzil} is the baseline model finetuned on Tanzil domain data. \textbf{All-FT} is the baseline finetuned model on EMEA, Tanzil and Tech domain data. \textbf{DoSS} is our proposed model adapted to EMEA, Tanzil and Tech domains.}
    \label{tab:finetuning}
\end{table}

To assess the effect of DoSS hyper-parameters $\alpha$ and $\beta$ which specify the percentage of encoder and decoder parameters that DoSS was not allowed to modify, we experimented with applying DoSS on three domains: medical, religion, and tech. We experimented with $\alpha$ and $\beta$ values of 0.4,0.5,0.6,0.8,0.9. Table \ref{tab:alphabeta} shows that we obtained the best performance with $\alpha=0.6$ and $\beta=0.6$ and that the worst BLEU corresponds to the case where only 10\% of  encoders parameters were allowed to change per domain. $\alpha$ shows stronger correlation ($\rho=-0.74$) with the model performance on average for all three domains that align with the hypothesis that encoder needs more domain-specific information but decoder might have a weaker correlation with model performance ($\rho=-0.54$). We hypothesize that decoder needs less domain-specific parameters due to the inherited domain-specific information represented by the encoder.

Moreover we find that as the domain dataset size increases the more decoder parameters need to be allowed to change (lower $\beta$s are needed for larger datasets). Intuitively we attribute that to the model's need to adapt the decoder to more domain-specific terms as the domain dataset size increases.

\begin{table}[ht!]
    \centering
    \small
    \begin{tabularx}\linewidth{X X X X X | X }
        \hline
        $\alpha$ & $\beta$ & EMEA & Tanzil & Tech & Average \\
        \hline
        0.6 & 0.6 & \textbf{54.03} & 18.59 & 56.04 & \textbf{42.89}\\
        0.7 & 0.7 & 52.38 & 18.65 & 57.17 & 42.73\\
        0.8 & 0.8 & 51.46 & 18.33 & 55.76 & 41.85\\
        0.9 & 0.9 & 48.46 & 18.61 & 47.39 & 38.16\\
        0.4 & 0.6 & 52.24 & 18.41 & 57.39 & 42.68\\
        0.5 & 0.6 & 53.10 & 18.53 & 56.21 & 42.61\\
        0.6 & 0.8 & 52.12 & \textbf{18.82} & 57.23 & 42.72\\
        0.6 & 0.9 & 51.27 & 18.70 & \textbf{58.36} & 42.78\\
        \hline
    \end{tabularx}
    \caption{Effect of $\alpha$ and $\beta$ on BLEU}
    \label{tab:alphabeta}
\end{table}

\subsection{Domain Extensibility}
One of the main advantages of DoSS is the ability to adapt existing models to new domains, without dramatic drops in the performance of existing domain(s) and also with maintaining competitive performance to domain-specific fine-tuning on the domain-to-add. 

We conduct three experiments to examine the effect of different masking schemes and/or whether or not we train on the domain-to-add data only or re-use the existing domain data in addition to the domain-to-add. 
\begin{itemize}   
    \item We construct the mask without any constraints and continue training only on the domain-to-add data.
    \item We construct the mask without any constraint and continue training all pre-existing domains using all available domain data in addition to training data of the domain-to-add.
    \item We construct the mask with constraint to be disjoint from the union of all existing domain masks and continue training only on the domain-to-add data.
\end{itemize}
  
  In all of these we keep the same experimental setup (EMEA, Tanzil, Tech) and try to add the legal domain using the JRC Aquis dataset. Table \ref{tab:domain_ext} shows multiple baselines (Namely: Zero-shot using the baseline model, Fine-tuning the baseline, Zero-shot using the DoSS model with an all 1s mask, Fine-tuning the DoSS model using an all 1s mask) as well as the results of the three previously mentioned main experiments.
  
\begin{table}[h!]
    \centering
    \small
    \begin{tabularx}\linewidth{m{6.5em} m{1.75em} m{1.35em} m{1.35em} m{1.7em} | m{1.7em} m{1.4em}}
        \hline
            & EMEA & Tanzil & Tech & JRC & AVG & N.P.\\
        \hline
        Baseline & 41.52 & 16.70 & 33.00 & 33.61 & 31.20 & 0\\
        All-FT & 53.26 & \textbf{19.00} & 52.01 & 40.05 & 41.08 & 270\\
        DoSS & \textbf{54.03} & 18.59 & 56.04 & 22.25 & 37.73 & 0\\
        DoSS-FT & 49.36 & 11.40 & 41.79 & 41.37 & 35.98 & 270\\
        DoSS-JRC & 48.85 & 11.58 & 43.27 & 41.28 & 36.25 & 107\\
        DoSS-all-masks & 53.47 & 18.55 & \textbf{57.20} & 41.32 & \textbf{42.64} & 146\\
        DoSS-JRC-disjoint & 54.00 & 18.60 & 56.01 & \textbf{41.80} & 42.60 & 37\\
        \hline
    \end{tabularx}
    \caption{SacreBLEU scores for domain extension. N.P denotes the number of trainable parameters in Millions. \textbf{Baseline} is the general model trained on WMT19. \textbf{All-FT} is the baseline finetuned model on EMEA, Tanzil and Tech domain data. \textbf{DoSS} is our proposed model adapted to EMEA, Tanzil and Tech domains. \textbf{DoSS-FT} is the DoSS finetuned model on JRC domain data only. \textbf{DoSS-JRC} is the continuation of applying DoSS on JRC domain only. \textbf{DoSS-all-masks} is the continuation of applying DoSS on EMEA, JRC, Tanzil and Tech domains. \textbf{DoSS-JRC-disjoint} is the continuation of applying DoSS on JRC domain only using disjoint mask.
    }
    \label{tab:domain_ext}
\end{table}

We observe that fine-tuning the DoSS model without any mask (a mask of all 1s) outperforms fine-tuning the original baseline model. In both cases we observe significant regressions on pre-existing domains, however DoSS still maintains a marginally better performance across pre-existing domains than the fine-tuned baseline model.
The first experimental setup to generate an unconstrained new mask and train on JRC data only manages to maintain the model performance on JRC in comparison to directly fine-tuning the DoSS model while slightly mitigating observed regressions on pre-existing domains by 0.4 BLEU points on average. The second method of continue training on pre-existing domains while adding the new domain manages to improve pre-existing domains by 0.19 BLEU points recovering from a 8.31 BLEU points regression on average while improving JRC performance by 0.1 BLEU points. The final setup manages to completely preserve pre-existing domains performance which is expected since the domain-to-add mask is disjoint from pre-existing masks while also improving JRC performance by 0.5 BLEU points in comparison to the second method. The disjoint mask method has the advantage of quicker convergence since we train a fewer number of parameters using a smaller dataset (domain-to-add data only).

\section{Conclusion}
\label{conclusion}
In this paper, we propose a new efficient method for multi-domain adaptation by learning domain-specific sub-network (DoSS). DoSS can efficiently generalize to new domains  while preserving the performance of existing domains. For our experiments on de-en machine translation DoSS outperforms the strong baseline of continue training on multi-domain (medical, tech, religion) data by 1.47 BLEU points. Also for the interesting scenario of extension to new domains it outperforms continue training on multi-domain data (medical, tech, religion, legal) by 1.52 BLEU points.

In future work we plan to explore adding more domains, using domain classifiers during inference, experimenting with multi-lingual and multi-domain setup and looking into new ways of defining constrained masks. We could also explore applying the method on sparse architectures.

\bibliography{anthology,custom}
\bibliographystyle{acl_natbib}

\end{document}